\documentclass[conference]{IEEEtran}
\IEEEoverridecommandlockouts
\usepackage{cite}
\usepackage{amsmath,amssymb,amsfonts}
\usepackage[english]{babel}
\usepackage{algorithmic}
\usepackage{graphicx}
\usepackage{textcomp}
\usepackage{hyperref}
\usepackage{balance}

\usepackage{booktabs}
\usepackage{multirow}

\addto\extrasenglish{%
  
}

\usepackage[T1]{fontenc}
\usepackage{comment}
\usepackage{float} 
\usepackage{xcolor}
\usepackage{acronym}
\usepackage{bm}

\usepackage{xcolor}
\def\BibTeX{{\rm B\kern-.05em{\sc i\kern-.025em b}\kern-.08em
    T\kern-.1667em\lower.7ex\hbox{E}\kern-.125emX}}

\begin{document}

\title{Lesion-Aware Generative Artificial Intelligence for Virtual Contrast-Enhanced Mammography in Breast Cancer}

\author{

\IEEEauthorblockN{
Aurora Rofena\IEEEauthorrefmark{1},
Arianna Manchia\IEEEauthorrefmark{1},
Claudia Lucia Piccolo\IEEEauthorrefmark{2},
Bruno Beomonte Zobel\IEEEauthorrefmark{2}\IEEEauthorrefmark{3},
Paolo Soda\IEEEauthorrefmark{1}\IEEEauthorrefmark{4},
Valerio Guarrasi\IEEEauthorrefmark{1}
}

\IEEEauthorblockA{
\IEEEauthorrefmark{1} Unit of Computer Systems and Bioinformatics, Department of Engineering, \\
University Campus Bio-Medico of Rome, Italy \\
\{aurora.rofena, valerio.guarrasi, p.soda\}@unicampus.it}

\IEEEauthorblockA{
\IEEEauthorrefmark{2}
Department of Radiology, Fondazione Policlinico Campus Bio-Medico, Italy}

\IEEEauthorblockA{
\IEEEauthorrefmark{3}
Department of Radiology, Università Campus Bio-Medico di Roma, Italy}

\IEEEauthorblockA{
\IEEEauthorrefmark{4}
Department of Radiation Sciences, Radiation Physics, Biomedical Engineering,
Umeå University, Sweden \\
\{paolo.soda@umu.se\}}

}

\maketitle      
\begin{abstract}
Contrast-Enhanced Spectral Mammography (CESM) is a dual-energy mammographic technique that improves lesion visibility through the administration of an iodinated contrast agent. It acquires both a low-energy image, comparable to standard mammography, and a high-energy image, which are then combined to produce a dual-energy subtracted image highlighting lesion contrast enhancement. While CESM offers superior diagnostic accuracy compared to standard mammography, its use entails higher radiation exposure and potential side effects associated with the contrast medium. To address these limitations, we propose Seg-CycleGAN, a generative deep learning framework for Virtual Contrast Enhancement in CESM. The model synthesizes high-fidelity dual-energy subtracted images from low-energy images, leveraging lesion segmentation maps to guide the generative process and improve lesion reconstruction. Building upon the standard CycleGAN architecture, Seg-CycleGAN introduces localized loss terms focused on lesion areas, enhancing the synthesis of diagnostically relevant regions. Experiments on the CESM@UCBM dataset demonstrate that Seg-CycleGAN outperforms the baseline in terms of PSNR and SSIM, while maintaining competitive MSE and VIF. Qualitative evaluations further confirm improved lesion fidelity in the generated images. These results suggest that segmentation-aware generative models offer a viable pathway toward contrast-free CESM alternatives. 
\end{abstract}


\section{Introduction}

Contrast-Enhanced Spectral Mammography (CESM)
\cite{jochelson2021contrast} is a dual-energy mammographic imaging technique classified within level II breast diagnostic procedures.
Unlike standard Full Field Digital Mammography (FFDM), CESM involves the intravenous administration of an iodinated contrast agent to highlight hypervascularized regions, thereby enhancing diagnostic accuracy, particularly in patients with dense breast tissue.
The examination consists of acquiring both low-energy (LE) images—comparable to FFDM—and high-energy (HE) images, which are uninterpretable in isolation.
These images are subsequently processed to produce dual-energy subtracted (DES) images, in which parenchymal tissue is suppressed, thereby accentuating areas of contrast uptake and improving lesion visibility.
For diagnostic interpretation, radiologists analyze both LE and DES images to ensure comprehensive assessment.
CESM is proposed as a viable alternative to contrast-enhanced breast magnetic resonance imaging, offering comparable sensitivity, greater specificity due to less background enhancement, and improved speed, cost-effectiveness, and patient tolerability.
Despite its benefits, the widespread adoption of CESM may be limited by two main issues.
First, the use of iodinated contrast media poses a risk of adverse reactions, ranging from mild hypersensitivity to more severe complications, such as contrast-induced nephropathy, shortness of breath, or facial swelling. Second, CESM entails a higher radiation dose than FFDM, attributable to the dual-energy exposure necessary for acquiring HE images \cite{patel2018contrast}.
Consequently, there is growing scientific interest in developing alternative strategies that mitigate the risks associated with contrast agents and reduce radiation exposure while preserving the diagnostic efficacy of CESM. 
In this context, it is increasingly recognized that a comprehensive and multimodal view of the patient can significantly improve overall diagnostic performance~\cite{guarrasi2025systematic, francesconi2025class, ruffini2024multi, di2025graph, guarrasi2023multi, mogensen2025optimized, guarrasi2022optimized}. The fusion of anatomical and functional information supports more informed clinical decision-making, particularly in complex cases or patients with dense breast tissue. Therefore, replicating the diagnostic value of CESM through synthetic, contrast-free alternatives can contribute not only to patient safety but also to maintaining a holistic and accurate diagnostic framework.

Recent advancements in Generative Artificial Intelligence (Gen-AI) have paved the way for innovative approaches, particularly through the application of Deep Learning (DL) models such as Generative Adversarial Networks (GANs).
These models have been explored for Virtual Contrast Enhancement (VCE) in CESM, aiming to perform image-to-image translation by converting LE images into virtual DES images that emulate contrast enhancement \cite{jiang2021synthesis, rofena2024deep, rofena2025augmented}.
Given the similarity between LE and FFDM images, this approach holds the potential to generate DES-equivalent images without the need for contrast administration or additional radiation exposure beyond FFDM levels, thereby offering a safer and more patient-friendly alternative for breast lesion detection and characterization.
Within this context, this work pursues the following objectives:
\begin{itemize}
    \item developing a DL model for the VCE task in CESM, capable of generating realistic and accurate DES images from LE images;
    \item integrating tumor lesion segmentation maps into the training process to enhance lesion generation, potentially improving clinical applicability and diagnostic accuracy.
\end{itemize}

The remainder of this paper is organized as follows: 
Section~\ref{sec:Background} describes the state-of-the-art of Gen-AI for VCE in the CESM field; Section~\ref{sec:Materials} presents the data used; Section~\ref{sec:Methods} elaborates on the proposed methodology and experimental setup; Section~\ref{sec:Results} presents the findings and discusses their implications; finally, Section~\ref{sec:Conclusion} summarizes the key insights and practical implications of the study.

\section{Background} \label{sec:Background}
VCE has been recently explored in the CESM context.
Jiang et al. \cite{jiang2021synthesis} proposed a cGAN-based Synthesis Network that leverages LE images to synthesize corresponding DES images.
They employed a cycle-consistent approach to minimize information loss when converting from high to low tissue contrast images.
Their method processes craniocaudal and mediolateral oblique views separately while introducing concatenation layers to enable dual-view information fusion.
Rofena et al. \cite{rofena2024deep} proposed a comparative study of three Gen-AI models, i.e., autoencoder, Pix2Pix \cite{pix2pix} and CycleGAN \cite{cyclegan} for translating LE images into DES images.
The study, supported by both quantitative and qualitative evaluations conducted with radiologists, demonstrated that CycleGAN has the potential to produce synthetic DES images that closely resembled the corresponding target DES images. 
The same authors further investigated the VCE task in the context of virtual biopsy \cite{rofena2025augmented}.
Exploiting the similarity between LE images and FFDM, they proposed a multimodal, multi-view virtual biopsy task that combines FFDM (i.e., LE images) with CESM (i.e., DES) in both craniocaudal and mediolateral oblique views.
Since clinical protocols do not include CESM for all patients, they employed CycleGAN to generate synthetic CESM images when missing.
Their results showed that integrating FFDM with synthetic CESM enhances virtual biopsy performance compared to using FFDM alone, highlighting the potential of VCE in the CESM setting.
However, to the best of our knowledge, no studies in this context have focused on generating DES images from LE images, using specific mechanisms for accurate lesion reconstruction. 
Addressing this gap could further refine the synthetic image quality and improve downstream clinical applications, reinforcing the role of generative AI in enhancing CESM-based diagnostics.

\section{Materials}  \label{sec:Materials}

\begin{figure}
    \centering
    \includegraphics[width=1\linewidth]{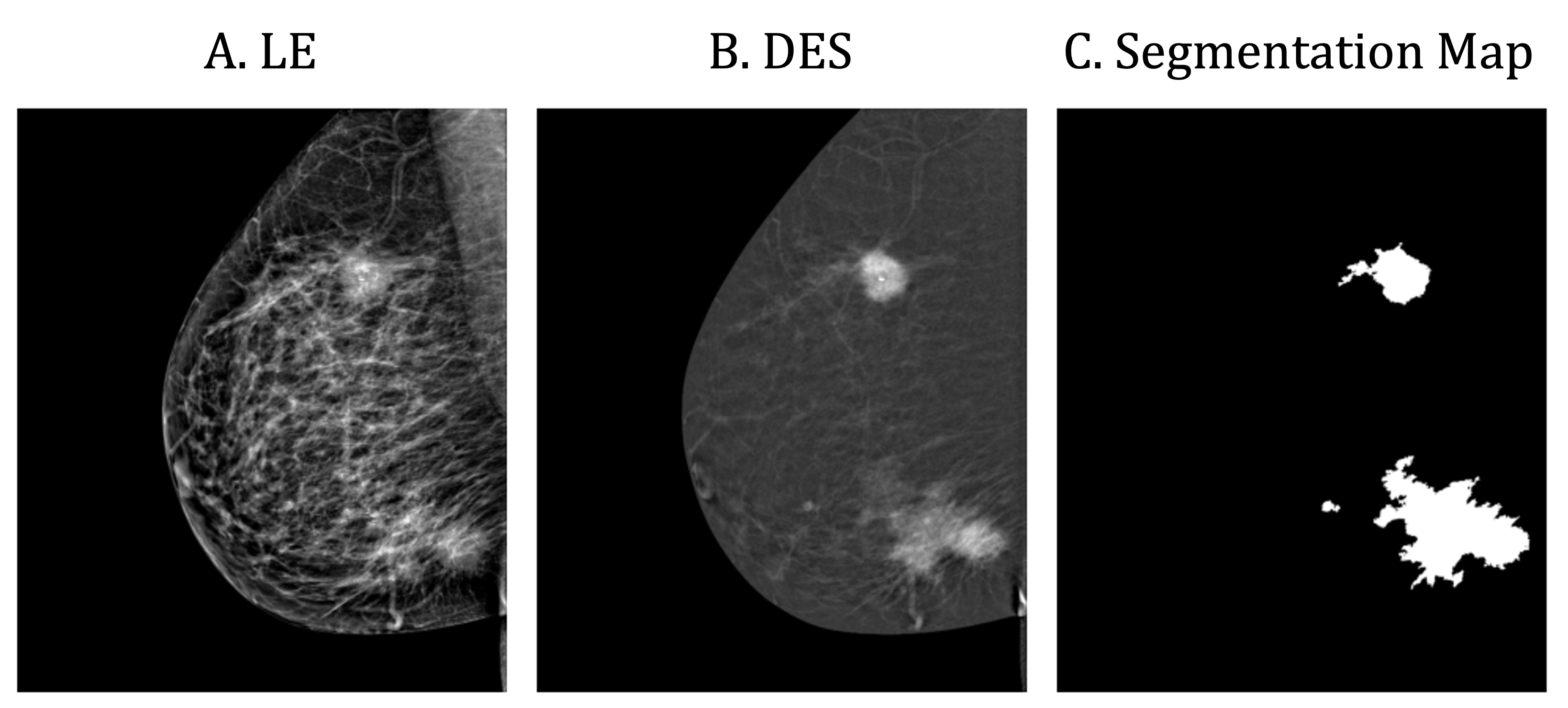}
    \caption{(A) LE image, (B) DES image, (C) Segmentation Map.}
    \label{fig:materials}
\end{figure}

In this study, we employ the CESM@UCBM dataset \cite{rofena2025augmented}, which comprises CESM examinations from 204 patients aged 31 to 90 years, with an average age of 56.7 years and a standard deviation of 11.2 years.
All examinations were performed at the Breast Unit of the Fondazione Policlinico Universitario (FPUCBM) Campus Bio-Medico in Rome between September 29, 2021, and August 23, 2023, using the Senographe Pristina full-field digital mammography system (GE Healthcare).
The dataset includes a total of 2,278 images in DICOMDIR format, equally distributed between 1,139 LE images and 1,139 DES images.
Of these, 1998 images have a resolution of $2850 \times 2396$ pixels, while the remaining 280 images are $2294 \times 1916$ pixels.
According to medical reports, 117 patients were diagnosed with at least one tumor lesion.
Lesion segmentation was performed on the DES images by radiologists at FPUCBM using 3D Slicer software, resulting in 425 segmentation maps.
Since LE and DES images are spatially aligned, the obtained segmentations can be directly used for both modalities.
Figure \ref{fig:materials} shows an example of a pair of LE and DES images with the corresponding segmentation map.

\section{Methods}  \label{sec:Methods}

\begin{figure*}[t]
    \centering    \includegraphics[width=1\linewidth]{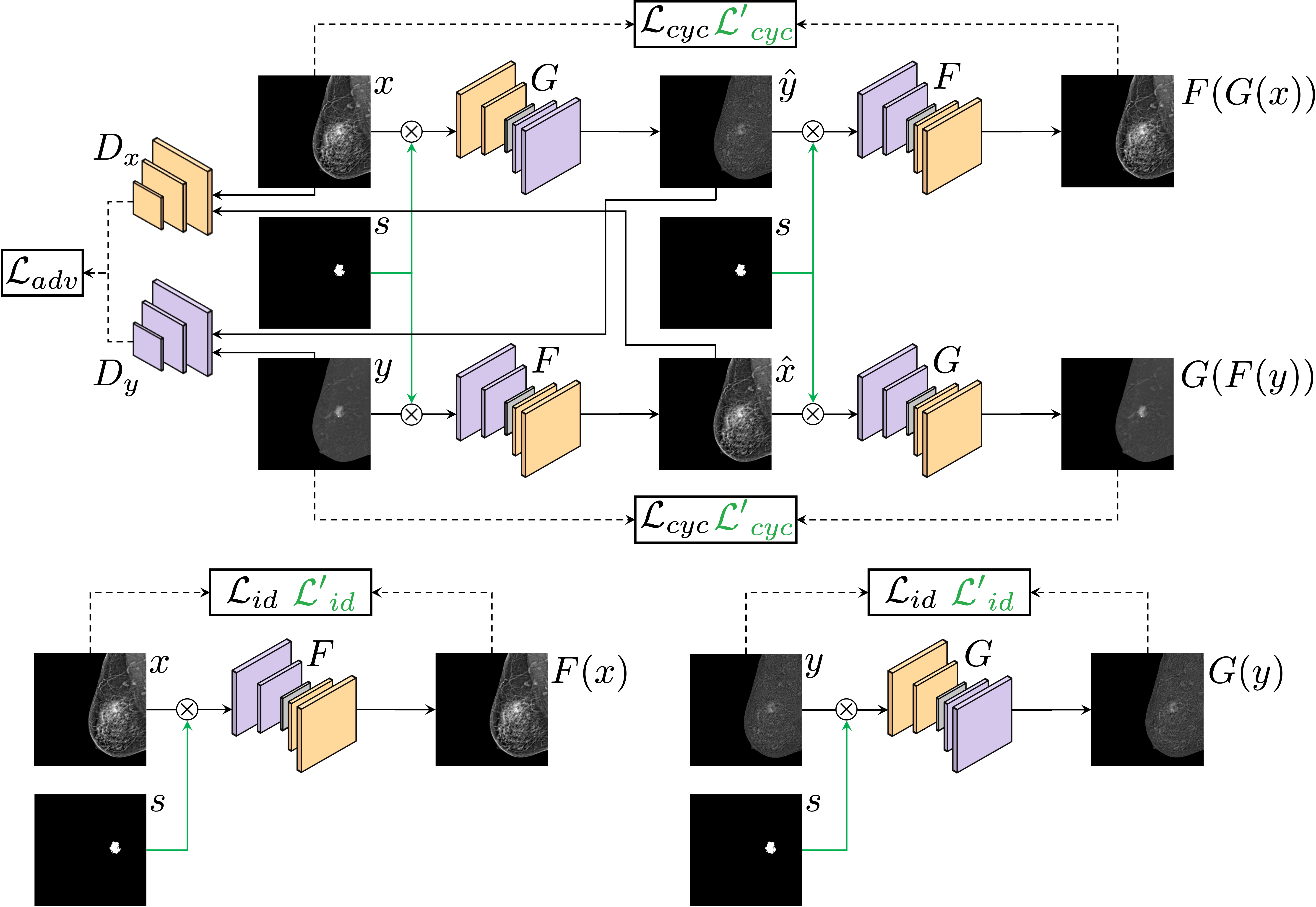}
    \caption{Schematic representation of the proposed method. We propose Seg-CycleGAN, a model based on the CycleGAN architecture designed to generate DES images from LE images. Unlike the standard CycleGAN, Seg-CycleGAN leverages tumor lesion segmentation maps during training to guide the image translation process.}
    \label{fig:method}
\end{figure*}

\subsection{Generative Model}
Building upon prior research \cite{rofena2024deep, rofena2025augmented}, we investigated the CycleGAN architecture to address the VCE task on CESM, which involves translating LE images into their corresponding DES images.
To enhance the quality of lesion generation, we extended the standard CycleGAN by introducing additional loss terms that leverage segmentation maps.
These terms selectively focus on lesion regions, guiding the model to more accurately reconstruct lesion-specific features.
A schematic representation of the proposed method is provided in Figure \ref{fig:method}.
CycleGAN~\cite{cyclegan} is a type of GAN that translates images between two domains without requiring paired samples.
It consists of two generators, $G$ and $F$, and two discriminators, $D_y$ and $D_x$.
The generator $G$ maps the $X$ domain to the $Y$ domain, while $F$ performs the inverse mapping.
The discriminator $D_y$ distinguishes between real images $y$ in domain $Y$ and generated images $\hat{y}~=~G(x)$, whereas $D_x$ differentiates real images $x$ from generated images $\hat{x}~=~F(y)$.
Adversarial training employs the adversarial loss $\mathcal{L}_{adv}(G,D_y)$, wherein $G$ aims to maximize the loss to fool $D_y$, while $D_y$ aims to minimize it, thereby improving its ability to distinguish between real and generated samples. $\mathcal{L}_{adv}(G,D_y)$ is defined as follows:
\begin{equation}
   \mathcal{L}_{adv}(G,D_y) = \mathbb{E}_{y}[\log(D_y(y))]+\mathbb{E}_{x}[\log(1-D_y(G(x))]
   \label{eq:cg_lgan}
\end{equation}
Similarly, the adversarial loss $\mathcal{L}_{adv}(F, D_x)$ governs the adversarial training between $F$ and $D_x$.
To enforce consistent mappings, the generators employ the cycle consistency loss $\mathcal{L}_{cyc}(G, F)$, ensuring that an input image can be reconstructed after forward transformations, i.e., $x \rightarrow G(x) \rightarrow F(G(x)) \approx x$, and backward transformation, i.e., $y \rightarrow F(y) \rightarrow G(F(y)) \approx y$.
This loss minimizes the difference between original images, i.e., $x$ and $y$, and the corresponding recovered images, i.e., $F(G(x))$ and $G(F(y))$, and is defined as:
\begin{equation}
   \mathcal{L}_{cyc}(G,F) = \mathbb{E}_{y}[||G(F(y)-y||_1]+\mathbb{E}_{x}[||F(G(x))-x||_1]
   \label{eq:cg_lcyc_1}
\end{equation}
Finally, an identity mapping loss $\mathcal{L}_{id}(G,F)$ ensures that when samples from the target domain are input into the generator, the output remains unchanged. It is formulated as:
\begin{equation}
   \mathcal{L}_{id}(G,F) = \mathbb{E}_{y}[||G(y)-y||_1]+\mathbb{E}_{x}[||F(x)-x||_1]
   \label{eq:cg_lcyc_2}
\end{equation}
Thus, the full CycleGAN objective is:
\begin{equation}
\begin{aligned}
   G^*, F^* &= \arg \min_{G, F} \max_{D_x, D_y} \bigl[ \mathcal{L}_{adv}(G, D_y) + \mathcal{L}_{adv}(F, D_x) \\
   &\quad + \lambda_1 \mathcal{L}_{cyc}(G, F) + \lambda_2 \mathcal{L}_{id}(G, F) \bigr]
\end{aligned}
\label{eq:cg_objective}
\end{equation}
where $\lambda_1$ and $\lambda_2$ balance the respective contributions of each loss term.

To improve lesion generation in the VCE task, we integrated segmentation maps $s$ into the training process.
By leveraging these maps, we introduced localized versions of the cycle consistency and identity mapping losses, $\mathcal{L'}_{cyc}(G, F)$ and $\mathcal{L'}_{id}(G,F)$, which focus exclusively on comparing the lesion areas of the generated and target images, rather than the entire images.
These localized losses are defined as follows:

\begin{equation}
\begin{aligned}
   \mathcal{L'}_{cyc}(G,F) &= \mathbb{E}_{y}\bigl[ \| s \cdot (G(F(y)) - y) \|_1 \bigr] \\
   &\quad + \mathbb{E}_{x}\bigl[ \| s \cdot (F(G(x)) - x) \|_1 \bigr]
   \label{eq:cg_lcyc_1}
\end{aligned}
\end{equation}

\begin{equation}
\begin{aligned}
   \mathcal{L'}_{id}(G,F) &= \mathbb{E}_{y}\bigl[ \| s \cdot (G(y) - y) \|_1 \bigr] \\
   &\quad + \mathbb{E}_{x}\bigl[ \| s \cdot (F(x) - x) \|_1 \bigr]
   \label{eq:cg_lcyc_2}
\end{aligned}
\end{equation}

We combined each localized loss term with its corresponding global counterpart and introduced a weighting factor $\gamma$ to balance their contributions to the overall objective function.
Consequently, the resulting loss function becomes:

\begin{equation}
\begin{aligned}
   G^*, F^* &= \arg \min_{G, F} \max_{D_x, D_y} \Bigl\{ \mathcal{L}_{adv}(G, D_y) + \mathcal{L}_{adv}(F, D_x) \\
   &\quad + \lambda_1 \bigl[ \mathcal{L}_{cyc}(G, F) + \gamma \cdot \mathcal{L'}_{cyc}(G, F) \bigr] \\
   &\quad + \lambda_2 \bigl[ \mathcal{L}_{id}(G, F) + \gamma \cdot \mathcal{L'}_{id}(G, F) \bigr] \Bigr\}
   \label{eq:new_cg_objective}
\end{aligned}
\end{equation}

This formulation encourages the generators to focus on accurately reconstructing lesion areas during training while preserving the overall image structure, thereby improving the quality of lesion generation in the VCE task.
To distinguish this approach from the standard CycleGAN trained with Equation \ref{eq:cg_objective}, we refer to our model as Seg-CycleGAN, which explicitly integrates lesion segmentation into the training objective.

\subsection{Experimental Set-up}
We pre-processed all the images of the CESM@UCBM dataset to ensure data consistency and uniformity.
The preprocessing pipeline consisted of four main steps: (i) zero-padding to convert images into square shapes, (ii) contrast stretching to adjust brightness levels, (iii) pixel value normalization to the $[0, 1]$ range, and (iv) resizing to 256 × 256 pixels.
To enhance generalization and prevent overfitting, we applied random data augmentation to the training set, including vertical and horizontal shifts (up to ±10\% of the original size), zoom variations (±10\%), horizontal flips, and rotations (up to ±15°).
Corresponding segmentation maps underwent the same augmentations.
Experiments were conducted using stratified 10-fold cross-validation, ensuring an equal distribution of samples with segmentation maps across folds.
The dataset was split into training (80\%), validation (10\%), and test (10\%) sets.
To facilitate training, we first pre-trained our model on a publicly available CESM dataset \cite{bib:PublicDataset}, applying the same preprocessing steps used for the CESM@UCBM dataset.
Since segmentation maps were unavailable in this dataset, we trained the model using the standard loss function (Equation \ref{eq:cg_objective}).
The resulting weights served as initialization for subsequent training on the CESM@UCBM dataset, where the loss function described in Equation \ref{eq:new_cg_objective} was employed, exploring different values for the $\gamma$ parameter.
We performed the training for a maximum of 200 epochs, with early stopping triggered if the validation loss did not improve for 50 consecutive epochs.
We used the Adam optimizer for both the generator and discriminator networks, with a learning rate of $10^{-5}$, weight decay of $10^{-5}$, beta of 0.5, and momentum of 0.999.
For loss functions, we employed the mean squared error for the adversarial losses $\mathcal{L}_{adv}(G, D_y)$ and $\mathcal{L}_{adv}(F, D_x)$.
We employed the L1 loss for both the cycle consistency losses, $\mathcal{L}_{cyc}(G, F)$ and $\mathcal{L'}_{cyc}(G, F)$, with a weighting factor of $\lambda_1 = 10$, as well as for the identity mapping losses, $\mathcal{L}_{id}(G, F)$ and $\mathcal{L'}_{id}(G, F)$, weighted by $\lambda_2 = 5$.

\subsection{Evaluation}
We assessed model performance using four quantitative metrics:
\begin{itemize}
    \item Mean Squared Error (MSE): measures mean squared difference between the pixel values of the target and the generated images. It varies in the range $[0,\infty]$; the lower its value, the higher the quality of the generated image.
    \item Peak Signal-to-Noise Ratio (PSNR): defined as the ratio of the maximum possible power of a signal to the power of the noise corrupting the signal. It is expressed in decibels and higher values indicate better quality.
    \item Visual Information Fidelity (VIF): evaluates the information preservation between test and target images. It varies in the range $[0,1]$ and the higher its value, the higher the quality of the synthetic image.
    \item Structural Similarity Index Measure (SSIM): evaluates the structural similarity between images. It varies in the range $[0,1]$ and the higher its value, the higher the similarity between the two images.
\end{itemize}

\section{Results and Discussion}  \label{sec:Results}

\begin{table*}[t]
\caption{Quantitative Evaluation.}
\label{tab:quantitative_results}
    \centering
    \resizebox{0.8\textwidth}{!}{%
    \begin{tabular}{l c c c c c}
        \toprule
        \textbf{Model} & $\gamma$ &  \textbf{MSE (↓)} ($10^{-2}$) & \textbf{PSNR (↑)} & \textbf{VIF (↑)} ($10^{-2}$) & \textbf{SSIM (↑)} ($10^{-2}$) \\
        \midrule
        \textbf{CycleGAN} & / & $0.40 \pm 0.11$ & $26.79 \pm 0.54$ & \textbf{17.84 $\pm$ 0.83} & $85.32 \pm 0.73$ \\
        \midrule
        \multirow{2}{*}{\textbf{Seg-CycleGAN}} 
        & 35  & $0.40 \pm 0.10$ & \textbf{26.88 $\pm$ 0.82} & $17.73 \pm 0.26$ & $85.35 \pm 0.74$ \\
        \cmidrule(lr){2-6} 
        & 100 & $0.40 \pm 0.10$ & $26.84 \pm 0.63$ & $17.46 \pm 0.67$ & \textbf{85.37 $\pm$ 0.79} \\
        \bottomrule
    \end{tabular}}
\end{table*}

\begin{figure}[t]
    \centering
\includegraphics[width=1\linewidth]{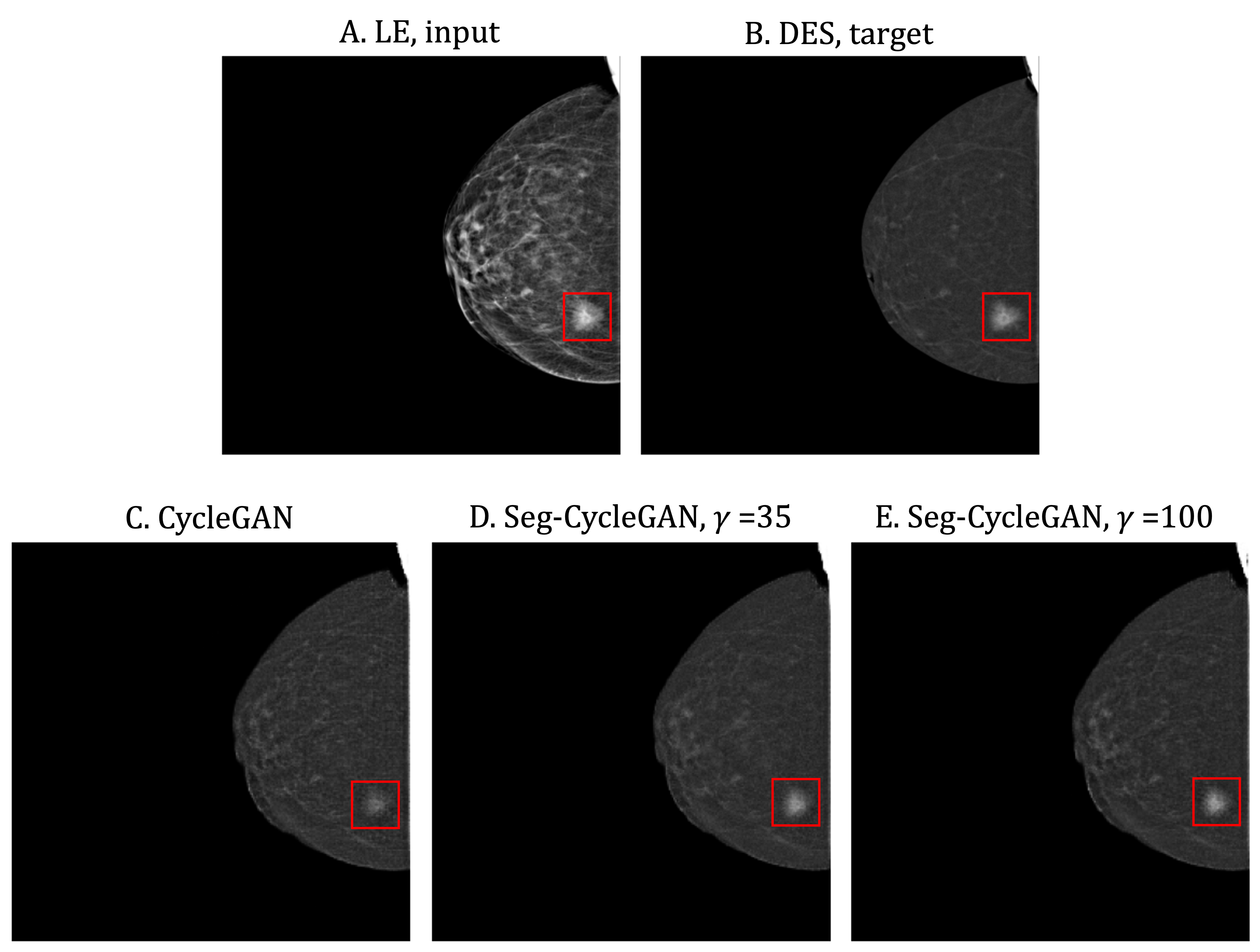}
    \caption{(A) LE input image, (B) DES target image, (C) CycleGAN output, (D) Seg-CycleGAN with $\gamma=35$ output, (E) Seg-CycleGAN with $\gamma=100$ output.}
    \label{fig:output}
\end{figure}

\begin{figure}[t]
    \centering
    \includegraphics[width=1\linewidth]{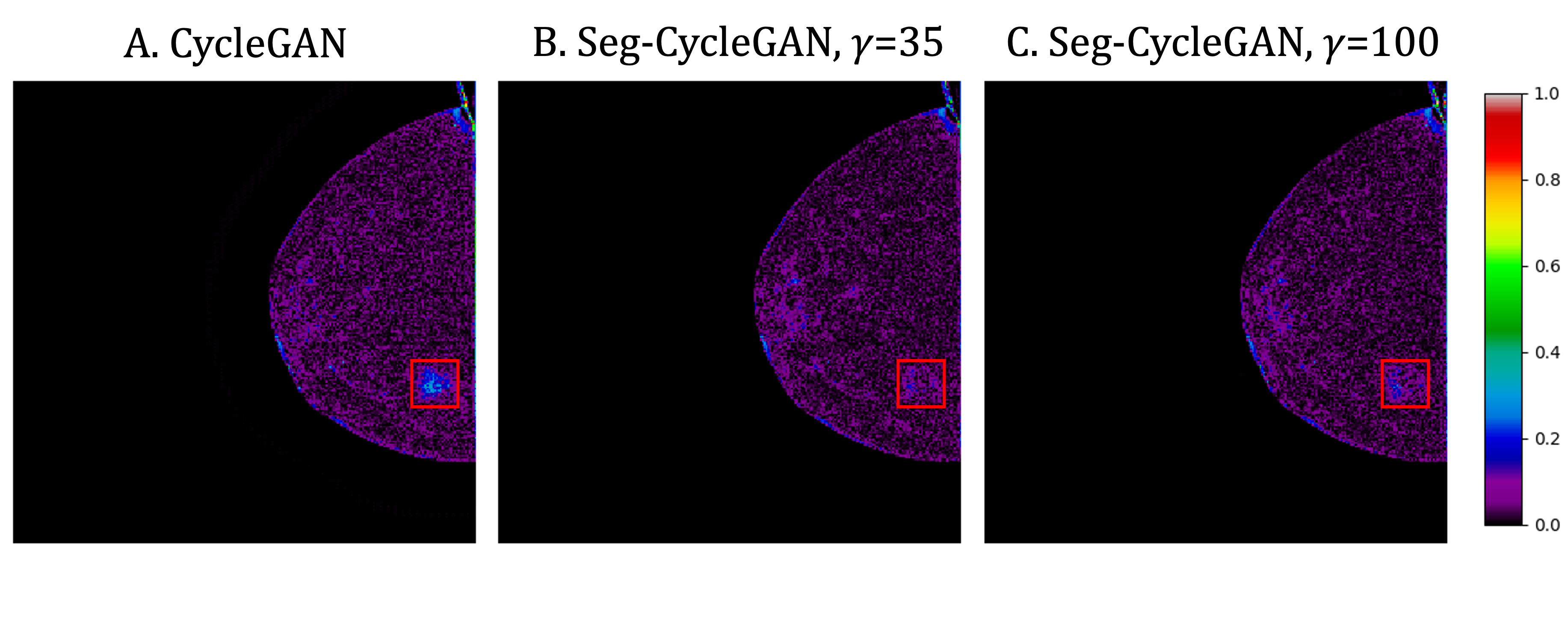}
    \caption{Heatmaps generated by comparing the target DES image and synthetic images generated by (A) CycleGAN, (B) Seg-CycleGAN with $\gamma=35$ (C), and Seg-CycleGAN with $\gamma=100$.}
    \label{fig:Heatmap}
\end{figure}

Table \ref{tab:quantitative_results} summarizes the quantitative evaluation results obtained by averaging performance metrics over the 10 folds, with results reported as mean ± standard deviation.
To establish a baseline, we first report the performance of the standard CycleGAN model, trained according to the loss function defined in Equation \ref{eq:cg_objective}.
In order to assess the impact of incorporating lesion-aware supervision, we compare this baseline with two configurations of the proposed Seg-CycleGAN model, using representative values of the segmentation-guided weighting parameter $\gamma$, specifically $\gamma = 35$ and $\gamma = 100$.
These values were selected based on empirical observations, given the absence of a clear deterministic trend linking performance to $\gamma$.
The results indicate that Seg-CycleGAN, for both $\gamma$ values, achieves higher average values for PSNR and SSIM compared to the standard CycleGAN, indicating an improved fidelity of the generated images.
Notably, the highest PSNR is observed with $\gamma =
35$, suggesting superior global similarity to the target image, while the highest SSIM, indicating perceptual and structural alignment, is achieved with $\gamma = 100$.
Although the standard CycleGAN yields the highest VIF score, the VIF values for both Seg-CycleGAN variants remain within an acceptable range, indicating that the informative content critical for diagnosis is effectively preserved.
MSE values remain comparable across all models, further supporting the stability of the generative process.
Figure \ref{fig:output} presents visual examples of an input image, the corresponding target image, and the outputs generated by the models. 
Specifically, the top section of the figure displays a pair of LE and DES images from the CESM@UCBM dataset, which serve as the input and target for the proposed models, respectively.
The presence of a malignant tumor mass, outlined by a red box, is observed in these images.
The bottom section of the figure illustrates, from left to right, the output generated by CycleGAN, followed by the outputs of Seg-CycleGAN with $\gamma = 35$ and $\gamma = 100$.
In each image, the region corresponding to the malignant tumor mass is consistently highlighted with a red box.
Upon examining the outputs, it can be observed that the background tissue is reconstructed with similar quality across all three models and remains consistent with the target image.
However, when focusing on the lesion region, the Seg-CycleGAN models demonstrate a more accurate reconstruction compared to CycleGAN.
This qualitative assessment is further reinforced by Figure \ref{fig:Heatmap}, which displays heatmaps illustrating the error distribution between the outputs of the three models and the target image.
In each heatmap, the region corresponding to the malignant tumor mass is delineated with a red box.
The heatmaps reveal that, while reconstruction errors in the background tissue are similar across all models, the error within the lesion area is substantially lower in the outputs of Seg-CycleGAN compared to CycleGAN.
This highlights the benefit of incorporating segmentation maps during training.

From a clinical perspective, these improvements have significant implications.
CESM images are used to detect and characterize suspicious lesions, especially in patients with dense breast tissue where FFDM may be insufficient.
By improving the quality and structural accuracy of synthesized DES images, particularly in regions corresponding to tumor tissues, Seg-CycleGAN may support more accurate computer-aided diagnosis systems or serve as a viable tool in scenarios where DES images are unavailable or incomplete.
This could ultimately enhance radiological workflows, reduce the need for repeat imaging, and assist in earlier and more precise detection of breast cancer.

\section{Conclusion}  \label{sec:Conclusion}
This study introduced Seg-CycleGAN, an enhanced version of CycleGAN tailored for VCE in CESM, aimed at synthesizing high-fidelity DES images from corresponding LE images.
By incorporating lesion segmentation maps into the training process, the model is explicitly guided to prioritize the accurate reconstruction of tumor regions.
Quantitative results show that Seg-CycleGAN outperforms the standard CycleGAN in terms of PSNR and SSIM, while maintaining comparable MSE and only a slight decrease in VIF. Qualitative analyses with heatmaps further confirm its improved ability to reconstruct lesion areas with greater visual fidelity, without compromising background consistency.
These findings suggest that incorporating segmentation-aware objectives enhances lesion-specific generation quality, offering a promising step toward safer and contrast-free alternatives to CESM.
However, in this study the segmentation maps only influence the generation of images by the generators but do not influence the decisions of the discriminators.
As a future direction, the architecture could be extended by incorporating lesion segmentation maps into the training process of the discriminators. This would encourage the models to focus more precisely on tumor regions when distinguishing real from synthetic images, potentially enhancing the realism and diagnostic relevance of the generated DES images. Furthermore, integrating this approach into multimodal diagnostic pipelines could support more comprehensive and patient-friendly breast cancer assessments. Emphasis will also be placed on the development of multimodal explainable AI strategies, enabling clinicians to better interpret model decisions across imaging modalities and improving trust and transparency in clinical applications~\cite{guarrasi2024multimodal}.

\section*{Acknowledgments}
Aurora Rofena is a Ph.D. student enrolled in the National Ph.D. in Artificial Intelligence, XXXVIII cycle, course on Health and life sciences, organized by Università Campus Bio-Medico di Roma. 
This work was partially founded by: i) PNRR MUR project PE0000013FAIR, ii) Cancerforskningsfonden Norrland project MP23-1122, iii) Kempe Foundation project JCSMK24-0094, iv) PNRR M6/C2 project PNRR-MCNT22023-12377755, v) Università Campus Bio-Medico di Roma under the program “University Strategic Projects” within the project “AI-powered Digital Twin for next-generation lung cancEr cAre (IDEA)”.

Resources are provided by the National Academic Infrastructure for Supercomputing in Sweden (NAISS) and the Swedish National Infrastructure for Computing (SNIC) at Alvis @ C3SE, partially funded by the Swedish Research Council through grant agreements no. 2022-06725 and no. 201805973.

\balance

\bibliographystyle{IEEEtran}
\bibliography{IEEEabrv, references.bib}

\end{document}